\documentclass[letterpaper]{article} 
\usepackage{aaai25}  
\usepackage{times}  
\usepackage{helvet}  
\usepackage{courier}  
\usepackage[hyphens]{url}  
\usepackage{graphicx} 
\usepackage{natbib}  
\usepackage{caption} 
\usepackage{algorithm}
\usepackage{algorithmic}

\usepackage{array}
\usepackage[caption=false,font=normalsize,labelfont=sf,textfont=sf]{subfig}
\usepackage{textcomp}
\usepackage{verbatim}
\usepackage{cite}

\usepackage{multirow}
\usepackage{amssymb}
\usepackage{bbding}
\usepackage{booktabs}
\usepackage{colortbl}

\usepackage[table,xcdraw]{xcolor}

\usepackage{pifont}

\usepackage{newfloat}
\usepackage{listings}
\DeclareCaptionStyle{ruled}{labelfont=normalfont,labelsep=colon,strut=off} 
\floatstyle{ruled}
\newfloat{listing}{tb}{lst}{}
\floatname{listing}{Listing}
\title{Seg2Box: 3D Object Detection by Point-Wise Semantics Supervision}
\author{
    Maoji Zheng\textsuperscript{\rm 1,\rm 2}, 
    Ziyu Xu\textsuperscript{\rm 1,\rm 2},
    Qiming Xia\textsuperscript{\rm 1,\rm 2},
    Hai Wu\textsuperscript{\rm 1,\rm 2}, 
    Chenglu Wen\textsuperscript{\rm 1,\rm 2}\footnote{Corresponding author.},  
    Cheng Wang\textsuperscript{\rm 1,\rm 2}
}
\affiliations{
    \textsuperscript{\rm 1}Fujian Key Laboratory of Sensing and Computing for Smart Cities, Xiamen University, China \\
    \textsuperscript{\rm 2}Key Laboratory of Multimedia Trusted Perception and Efficient Computing,  \\ 
    Ministry of Education of China, Xiamen University, China
    
    \{zhengmaoji, xuziyu, xiaqiming, wuhai\}@stu.xmu.edu.cn,
    \{clwen, cwang\}@xmu.edu.cn
}

\usepackage{bibentry}

\begin{document}

\maketitle

\begin{abstract}
LiDAR-based 3D object detection and semantic segmentation are critical tasks in 3D scene understanding. Traditional detection and segmentation methods supervise their models through bounding box labels and semantic mask labels. However, these two independent labels inherently contain significant redundancy. 
This paper aims to eliminate the redundancy by supervising 3D object detection using only semantic labels.
However, the challenge arises due to the incomplete geometry structure and boundary ambiguity of point-cloud instances, leading to inaccurate pseudo-labels and poor detection results. 
To address these challenges, we propose a novel method, named \textit{Seg2Box}.  
We first introduce a Multi-Frame Multi-Scale Clustering \textit{\textbf{(MFMS-C)}} module, which leverages the spatio-temporal consistency of point clouds to generate accurate box-level pseudo-labels. 
Additionally, the Semantic-Guiding Iterative-Mining Self-Training \textit{\textbf{(SGIM-ST)}} module is proposed to enhance the performance by progressively refining the pseudo-labels and mining the instances without generating pseudo-labels.
Experiments on the Waymo Open Dataset and nuScenes Dataset show that our method significantly outperforms other competitive methods by 23.7\% and 10.3\% in mAP, respectively. The results demonstrate the great label-efficient potential and advancement of our method.
\end{abstract}

%

\section{Introduction}
 
LiDAR-based 3D object detection and 3D semantic segmentation are widely applied in fields of autonomous driving~\cite{zhu2024spgroup3d, mao20233d}, robotics~\cite{robot-1, robot-2} and smart cities~\cite{3DODBS, Wang2019ASO}. 
Traditional 3D object detection and semantic segmentation frameworks rely on their unique labels to supervise learning processes. However, the annotations for semantic masks and bounding boxes inherently contain considerable redundancy, as they both implicitly convey the semantic content and geometric structure of instances. 

\begin{figure}[t]
  \centering
   \includegraphics[width=1.0\linewidth]{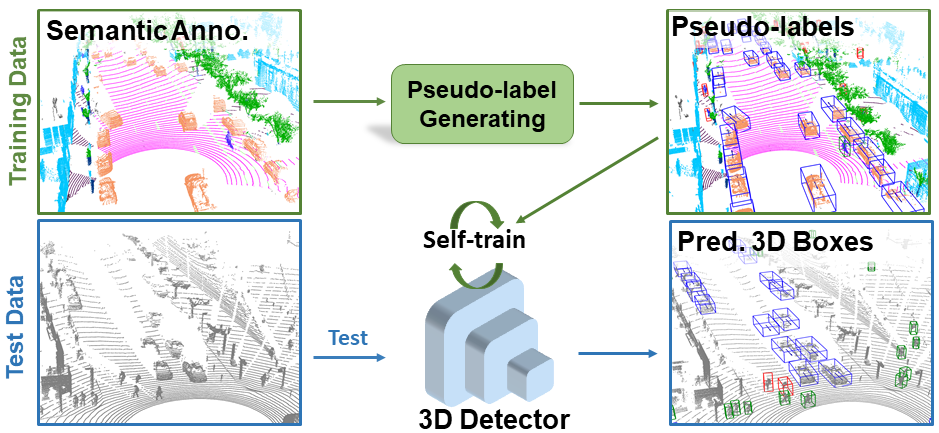}
   \caption{
   Our method uses only semantic annotation to train 3D object detection. It contains Pseudo-label Generation stage and Self-train Loop-improvement stage.
   }
   \label{fig:intro_four_pic}
\end{figure}

An intuitive solution to eliminate this redundancy is to use only bounding box labels and assign the semantic labels automatically to supervise the semantic segmentation task. Several studies have investigated the technique of semantics assignment.
For example, Box2Mask~\cite{box2mask} assigns semantics to each foreground point through instance clustering and box voting. Box2seg~\cite{box2seg} uses attention-based self-training and point class activation mapping to avoid using semantic labels. However, background information is missing in the bounding box annotation. In addition, bounding boxes mainly focus on a coarse level of instance layout. Using only bounding boxes to assign semantic labels inevitably introduces supervision noise, significantly decreasing the segmentation performance.

Compared to the coarse bounding box, the semantic label is more precise and detailed. 
Therefore, using only semantic labels possibly attains accurate detection and segmentation results and decreases label redundancy remarkably. However, the objects in 3D scenes are primarily sparse and self-occluded. Nowadays, how to recover the bounding boxes from the semantic labels to train the 3D detectors still remains a great challenge.

\begin{figure}[t]
  \centering
   \includegraphics[width=1.0\linewidth]{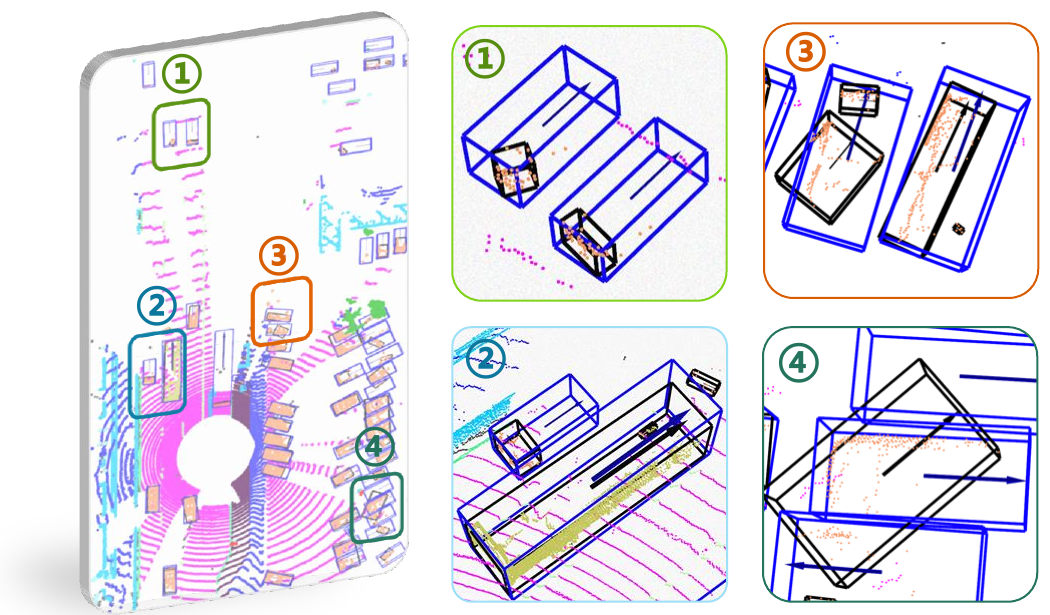}
   \caption{
   The challenges of generating box-level pseudo-labels from semantic labels. Blue boxes are the ground truth. And, black boxes are pseudo-labels generated by direct DBSCAN clustering. Points with different colors indicate different categories of objects. \ding{172},  \ding{173}: Incomplete objects due to sparse point cloud and occlusion. \ding{174}: Clustering one instance into multiple due to the truncated object. \ding{175}: Clustering multiple instances into one due to adjacent objects.
   }
   \label{fig:difficult}
\end{figure}

This paper mainly investigates two challenges of semantic label-supervised 3D object detection. (1) The incomplete geometry structure resulting from sparse point clouds (Fig. \ref{fig:difficult} - \ding{172}) and occlusion (Fig. \ref{fig:difficult} - \ding{173}) leads to pseudo-labels with erroneous size and position. (2) The boundary ambiguity arising from the truncated objects (Fig. \ref{fig:difficult} - \ding{174}) and the adjacent objects (Fig. \ref{fig:difficult} - \ding{175}) results in pseudo-labels with false boundaries. 
Consequently, the detector trained with the low-quality pseudo-labels can not achieve desirable performance.

To address these challenges, we develop a two-stage 3D object detection method, named \textbf{Seg2Box}. In the first stage, we introduce a novel Multi-Frame Multi-Scale Clustering \textbf{(MFMS-C)} module to generate accurate pseudo-labels. MFMS-C adopts a density-based spatial clustering design and then fits bounding boxes to clustering results ~\cite{oyster, cpd}. 
The key idea is to select the best one from a set of pseudo-label candidates.
Specifically, MFMS-C first generates the pseudo-label candidates by clustering with different radii. MFMS-C then selects the best pseudo-label depending on a newly designed Meta Shape and Fitting Score (MSF-Score), which measures the pseudo-label quality from the aspect of completeness, distribution of points, and shape. 
In addition, the static instances in consecutive frames tend to be more complete, allowing the creation of accurate pseudo-labels with precise position and size. 
In the second stage, we develop a Semantic-Guiding Iterative-Mining Self-Training \textbf{(SGIM-ST)} module to enhance the detection performance. Specifically, We iteratively mine the miss annotated instances and refine the pseudo-labels using spatio-temporal consistency.

We validated our method on widely used Waymo Open Dataset (WOD)~\cite{waymo} and nuScenes Dataset~\cite{nuScenes}. Our method significantly outperforms baseline methods, achieving a 23.7\% improvement in mAP L2 on WOD. Remarkably, compared to fully supervised methods by bounding boxes, our approach reached close to 95\% accuracy in vehicle AP L2 with an IoU threshold of 0.5 while using only semantic labels as supervision.

The main contributions of this work include:
(1) We proposed the first method for 3D object detection that relies solely on semantic label supervision. This innovation eliminates the redundancy between bounding box labels and semantic labels, offering a viable approach to reducing human labeling costs. 
(2) We proposed \textbf{MFMS-C} module for pseudo-label generation which significantly improves the accuracy of pseudo-labels.
(3) We proposed the \textbf{SGIM-ST} module, which significantly enhances the detection performance by iteratively mining instances without pseudo-labels and refining pseudo-labels.

\section{Related Work}

\subsubsection{3D Object Detection from Point Cloud.}  
In 3D object detection, fully supervised methods~\cite{centerpoint, fsfd,casa, hanet} have been extensively researched and display outstanding performance. However, labor-intensive and time-consuming manual annotation limits their wide application.
Weakly supervised methods aim to reduce the annotation burden by annotating only selected frames or instances. They identify unlabeled instances through teacher-student frameworks~\cite{Wang20203DIoUMatchLI, Han2024DualPerspectiveKE} or feature-level instance mining~\cite{coin, hinted}. Recently, unsupervised methods have been explored for the 3D object detection~\cite{oyster, cpd}. 
Despite these weakly supervised and unsupervised strategies vastly reducing the annotation cost, they are hard to obtain optimal performance. 
The primary obstacle is the insufficiency of information. Our method introduces cross-task supervision to reduce the annotation cost while ensuring sufficient supervised information. 

\subsubsection{Cross-Task Supervision.} 
Cross-task supervision is a strategy that leverages shared knowledge across different tasks to train or enhance specific models.
In 2D vision, Boxsup~\cite{boxsup} employs box annotations to supervise the training of semantic segmentation models, effectively bridging the gap between 2D detection and segmentation. In 3D vision, Box2Mask~\cite{box2mask} uses bounding box voting and instance clustering to assign semantics to points inside boxes for semantic segmentation. However, it neglects background instances not labeled by the detection task. To address this issue, Box2Seg~\cite{box2seg} introduces additional subcloud-level tags to estimate background pseudo-labels but get poor performance. Moreover, bounding boxes mainly focus on a coarse level of instance layout. Using only bounding boxes to assign semantic labels inevitably introduces supervision noise. Therefore, our method leverages more precise and detailed semantic labels to supervise 3D object detection. 
It not only attains accurate detection results but also decreases the label redundancy remarkably. 

\subsubsection{Pseudo-Label Generation for 3D Object Detection.} 
Pseudo-label generation for 3D object detection is to estimate latent boxes for unlabeled data.
Recent label-efficient methods ~\cite{coin, cs-det, ss3d} generate pseudo-labels by a pre-trained model to supplement the supervision data.
WS3D~\cite{ws3d-acm} proposes an unsupervised 3D object proposal module to select high-confident boxes from 3D anchors. However, fixed-size anchors limit their effect and usage scenarios. Nowadays, unsupervised methods introduce the point-distribution-based strategy to generate pseudo labels. Modest~\cite{modest} distinguishes dynamic instances from multi-trip LIDAR sequences and then estimates the pseudo-labels by DBSCAN~\cite{dbscan} clustering. Prototype and unsupervised tracking have been introduced by oyster~\cite{oyster} and cpd~\cite{cpd} to refine the pseudo-labels. 

\section{Method}

\begin{figure*}[t]
  \centering
   \includegraphics[width=0.9\linewidth]{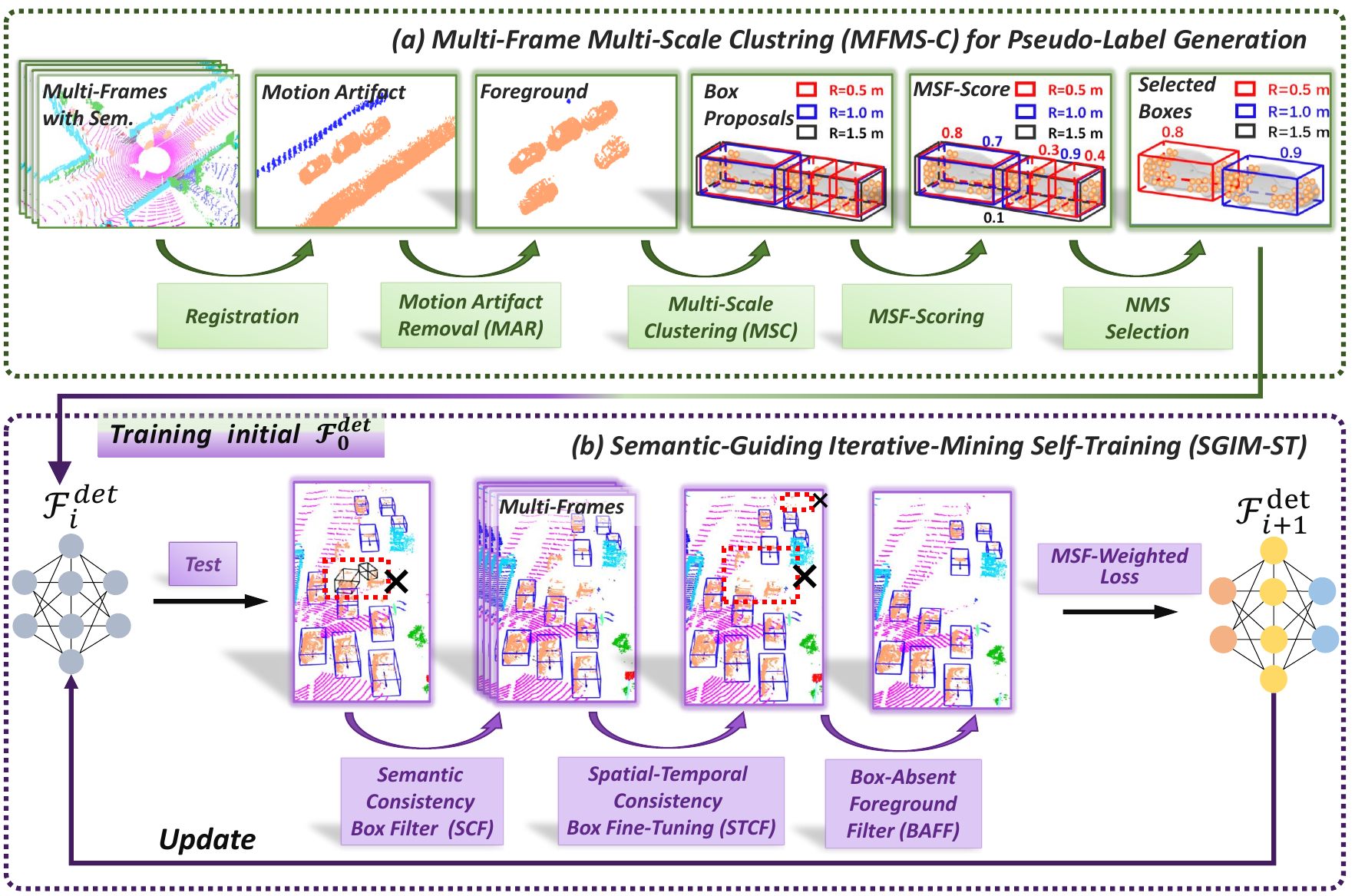}
   \caption{
    Illustration of Seg2Box framework. (a) MFMS-C generates box-level pseudo-labels from semantic points to train the initial detector $\mathcal{F}_{0}^{det}$. To address the challenges of pseudo-label generation due to incomplete geometry structure and boundary ambiguity, MFMS-C first generates numerous box proposals in consecutive frames using MSC. 
    After that, NMS Selection remains the high-quality proposals depending on MSF-Scoring which measures the quality of pseudo-labels. 
    (b) SGIM-ST enhances detection performance by iteratively mining the miss annotated instances
and refining the pseudo-labels through SCF, STCF, BAF, and MSF-Weighted Loss.
   }
   \label{fig:frame}
\end{figure*}

This paper introduces Seg2Box, a novel method for 3D object detection that uses only semantic labels as supervision. As illustrated in Fig.3, Seg2Box consists of two key stages: (1) Pseudo-labels generation by the MFMS-C module and (2) loop improvement by the SGIM-ST module. 

\subsection{Multi-Frame Multi-Scale Clustering for Pseudo-Label Generation}
Traditional methods ~\cite{oyster, dbscan} estimate box-level pseudo-labels from points by single radius-based DBSCAN~\cite{oyster, dbscan}.
However, these approaches fail to recover the labels of adjacent and truncated objects. The key reason is that the single radius-based clustering ignores the diversity of object density and location. 
In general, a small clustering radius is required for adjacent objects, while a large radius is needed for truncated or sparse objects.
How to accurately estimate the bounding boxes for all objects is still an unsolved challenge.          
We observe that clustering semantic points with multiple radii to construct a set of candidate boxes, and then selecting the best one by Non-Maximum Suppression(NMS), can significantly avoid false boundaries. Additionally, stationary objects in consecutive frames with multi-angle scans typically have more complete structures~\cite{cpd}.
Those observations motivate us to design MFMS-C module to generate accurate pseudo-labels. Specifically, MFMS-C consists of Motion Artifact Removal, Multi-scale Clustering and NMS-Selection. 

\subsubsection{Motion Artifact Removal.}
 Directly register continuous $2n+1$ frames \{$f_{-n},...,f_0,...,f_n$\} (i.e., past $n$, future $n$ frames, and current frame) to build dense point cloud $f_0^*$, resulting in artifacts from moving objects, which will affect the accuracy of pseudo-labels. To issue this problem, we first divide the visible area of $f_0^*$ into grids $G$ from the BEV perspective. Then, for each grid $G_{i,j}$, we count its maximal time $G_{i,j}^{ \;\,t}$ continuously occupied by foreground points. If $G_{i,j}^{ \;\,t}$ is less than a certain threshold $\varepsilon$ (related to the number of concatenated frames), $G_{i,j}$ is determined to be a moving area $A_{mov}$, otherwise, to be a static area $A_{sta}$. 
 We then keep all the foreground points in current frame $f_0$ and remove the points in moving areas $A_{mov}$ of other frames $f_{-n},...,f_{-1},f_1,...,f_n$ to found dense point cloud $f_0^*$. 

\subsubsection{Multi-Scale Clustering (MSC).} 
We follow the idea of density-based spatial clustering DBSCAN~\cite{dbscan}, bounding box fitting~\cite{box-fitting} on $f_0^*$ to generate a set of bounding boxes $\bar{b}$. However, traditional methods use a single radius for clustering, which causes mistaken instance division, resulting in false boundaries. To address this issue, we design our MSC module. For detail, we sample each radius $r$ from candidate radii ${CAND}_r$ to cluster $f_0^*$ and gain the initial proposals $\hat{b}_i$. After that, we concentrate all $\hat{b}_i$ to build bounding box candidates $\hat{\mathcal{B}}$ as the input of the next NMS-Selection module.

\subsubsection{Non-Maximum Suppression Selection (NMS-Selection).}
In the previous steps, we generate a lot of box candidates $\hat{\mathcal{B}}$ by MSC. However, each instance should only remain the best candidate at last. In object detection, NMS is widely used to eliminate redundant prediction boxes~\cite{NMS}. It suppresses the low-quality boxes through confident scores outputs by the network. Ground truth is needed to train the network, which is not available for our work.
In contrast, we introduce a newly designed Meta Shape and Fitting Score (MSF-Score) to evaluate the quality of pseudo-labels from the aspect of object completeness, distribution of points, and shape. Intuitively, good scoring should keep consistent with IoU scoring. As shown in Fig. \ref{score} (d), with increasing MSF-Score, the pseudo-label has a larger IoU with ground truth. 
MSF-Score consists of three sub-scores: Occupancy Score, Alignment Score, and  Meta Shape Score.  

\begin{figure}[t]
  \centering
   \includegraphics[width=0.9\linewidth]{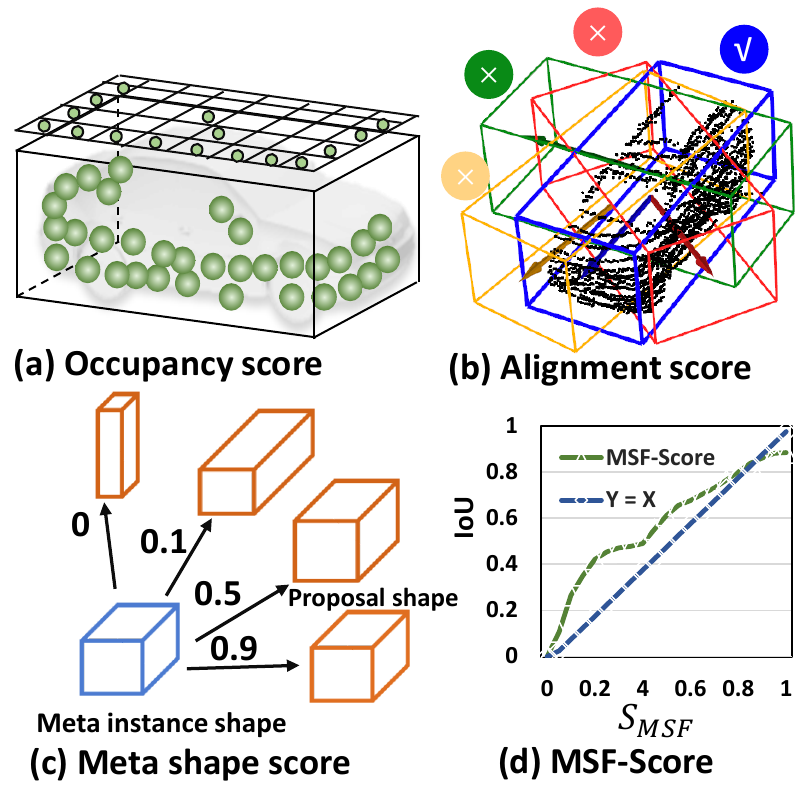}
   \caption{
        Meta Shape and Fitting Score (MSF-Score).
   }
   \label{fig:pseudo-labels-com}
\end{figure}

~\noindent {Occupancy (OCC) Score.} 
~Assuming the pseudo-label from a complete object is likely to be more accurate. As shown in Fig. \ref{score} - (a), we first divide the proposal box into grids with resolution $r \times r$ in BEV perspective~\cite{cpd}. The occupancy score indicates the proportion of grids occupied by foreground points. $S_{o}(b)$ is calculated as follow: 
\begin{equation}
S_{o}(b) = \frac{O}{r \times r}, \label{occ-score}
\end{equation}
where $O$ is the number of grids occupied by foreground points, and $r$ is the resolution.

~\noindent {Alignment (ALG) Score.}
~ Due to the nature of LiDAR scanning, points tend to be concentrated on the surface of instances.
Therefore, most of the points of the high-quality box should be close to the boundary (Fig. \ref{score} - (b)). 
We assume a box $b$ with angle $\alpha$. 
To evaluate $b$, we first look for the area with the highest point density in BEV perspective and fit a linear line for the points $P$ in this area, the angle of the line is $\theta$.
With an accurate pseudo-label, $\theta$ should align with $\alpha$ when points are concentrated along the length of the box, and it should be vertical to $\alpha$ when points are concentrated along the width of the box.
Motivate by this observation, we calculate the Alignment Score $S_{a}(b)$ as follow: 
\begin{equation}
S_{a}(b)=\left\{
\begin{array}{rcl}
1 - sin(|\alpha - \theta|), if |\alpha - \theta| < \frac{\pi}{2}  \\
1 - sin(|\alpha + \frac{\pi}{2} - \theta|), 
otherwise \\
\end{array} \right. 
\label{eq:sa}
\end{equation}

~\noindent {Meta Shape (MS) Score.}
~ In general, instances of the same category tend to have similar size ratios and fall within a certain size range (Fig. \ref{score} - (c)). For each class \textit{C}, we first construct its meta-shape $\mathcal{B} = \left\{l, w, h\right\}$, where $l$, $w$ and $h$ are the length, width and height, respectively. For these coarse-grained statistics, we can directly use category size information online~\cite{modest++}. Shape Score $S_{ms}(b)$ is calculated as follow:
\begin{equation}
S_{ms}(b)=\left\{
\begin{array}{rcl}
0 \ \ \ \ \ \ \ \ ,{b \leq 0.5 * \mathcal{B} \ or \ b \geq 2 * \mathcal{B}}\\
\min\left(0.05, \sum_{k} \mathcal{B}_k \log\left(\frac{\mathcal{B}_k}{b_k}\right)\right)  / {0.05}, \ \ else \\
\end{array} \right. 
\label{eq:ss}
\end{equation}
it indicates the ratio difference between $\mathcal{B}$ and $b$ ~\cite{cpd}, and $b$ is the proposal box size $\{\hat{l}, \hat{w}, \hat{h}\}$. 

Finally, We obtain the complete MFS-Score by linearly combining these three sub-scores:
\begin{equation}
MSF(b)=\left\{
\begin{array}{rcl}
\lambda_1 S_{o}(b) + \lambda_2 S_{a}(b) + \lambda_3 S_{ms}(b), \\ 
\lambda_1 + \lambda_2 + \lambda_3 = 1,\\
\end{array} \right. 
\label{score}
\end{equation}
where $\lambda_i$, $i = \{1,2,3\}$ control the weight of each sub-score. 

For each $b$ in box candidates $\hat{\mathcal{B}}$, we compute its MSF-Score $S_{\hat{\mathcal{B}}}^{MSF}$. Combined with NMS, we suppress the low-quality pseudo-labels and keep the best candidate $\hat{\mathcal{B}}_{final}$ with the highest MSF-Score to train the initial detection model.

\subsection{Semantic-Guiding Iterative-Mining Self-Training}
Self-training enhances the performance of the model by employing the output from the previous iteration as the input of the next iteration until the model converges.
Due to its ability to improve detection accuracy, self-training has been introduced to 3D object detection, especially for weakly supervised methods~\cite{ss3d, modest++, oyster}. However, there are still some issues that need to be considered in weakly supervised 3D detection self-training. (1) The instances without generating pseudo-label would be regarded as the background during training~\cite{WS3D}. (2) False classification (False positive) of similar objects is difficult to filter out by the traditional score-threshold method. (3) The detection results of instances of sparse scanning are often unreliable. (4) Despite refinement, some pseudo-labels are still inaccurate. These issues may mislead the model training process and accumulate with the increase of training iterations, which causes poor detection performance. 
To address these issues, we introduce our SGIM-ST module. SGIM-ST consists of four designs, which are described in detail below.

\subsubsection{Box-Absent Foreground Filter (BAF).} 
Foreground points of objects without pseudo-labels will be falsely regarded as background, which will mislead the model training. Since our supervision signal is semantic labels, foreground points and background points are already distinguishable. For each foreground point $p$, if $p$ is not inside any of the pseudo-labels, we filter it out from the point cloud. As self-training iterations progress, more and more precise pseudo-labels are mined, the filtered foreground points gradually decrease, and the scene becomes more and more complete.

\subsubsection{Semantic Consistency Box Filter (SCF).} False classification (False positive) of similar objects is a prevalent issue in 3D object detection. Traditional score-based filtering method often struggles to filter these errors. More critically, these errors can accumulate during iterative training, leading to reduced detection accuracy. For each predicted box $b^*$, we first extract the points $P_{in}$ inside $b^*$. We then filter out $b$ if the predicted category of $b$ is inconsistent with the semantics of $P_{in}$ or if multiple types of foreground semantic labels are found in $P_{in}$.

\subsubsection{Spatial-Temporal Consistency Box Fine-Tuning (STCF).} 
The quality of the predicted box decreases as the distance to the ego-car increases~\cite{mao20233d}. And, the imprecise pseudo-labels may mislead the training process. Our key idea is to broadcast the near-range high-quality predictions of static objects to other frames since stationary objects should be consistent between frames. We have already segmented the visual area into the static area $A_{sta}$ and the moving area $A_{mov}$ in the first stage. Then, the predicted boxes $\mathcal{B}$ in $A_{sta}$ are transformed into the global coordinate. After that, we use the NMS-Selection proposed in MSMF-C to score $\mathcal{B}$ and remain the best proposal $\mathcal{B}_{best}$ with the highest score. After that, we broadcast $\mathcal{B}_{best}$ back to a single frame by determining whether any foreground points in the single frame are inside $\mathcal{B}_{best}$. In this way, we refine the pseudo-labels for static objects even with spare points in some frames. We give up refining the moving objects since the poor tracking results.  
     
\subsubsection{MSF-Weighted Detection Loss (MSFLoss).} To suppress the false supervision of label noise caused by inaccurate pseudo-labels, we assign different training loss weights to instances according to pseudo-label quality. Formally, we calculate $\omega_{i}$ based on the MSF-Score $s^{msf}_{i}$ of pseudo-label:
\begin{equation}
\omega_{i}=\left\{
\begin{array}{rcl}
0, & & s^{msf}_{i} \le \theta_{L}\\
\frac{s^{msf}_{i} - \theta_{L}}{\theta_{H} - \theta_{L}}, & & \theta_{L} < s^{msf}_{i} < \theta_{H} \\
1, & & s^{msf}_{i} \ge \theta_{H}\\
\end{array} \right. 
\label{loss}
\end{equation}
where $\theta_{L}$ and $\theta_{H}$ are high-quality and low-quality thresholds respectively. The final MSF-weighed detection loss is calculated as follows:
\begin{equation}
\mathcal{L}_{MSF} = \frac{1}{N} \sum_{i=1}^{N} \omega_{i} (\mathcal{L}^{hm}_{i} + \mathcal{L}^{reg}_{i}),
\end{equation}
where N is the number of proposals, $\mathcal{L}^{hm}_{i}$, $\mathcal{L}^{reg}_{i}$ are heapmap loss and regression loss~\cite{centerpoint} between pseudo-labels and predictions.

\section{Experiments}

\begin{table*}[ht]
\setlength{\tabcolsep}{0.0mm}

\begin{tabular}{c|c|cccc|cccc|cccc}
\toprule
\rowcolor[HTML]{FFFFFF} 
\cellcolor[HTML]{FFFFFF}                         & \cellcolor[HTML]{FFFFFF}                     & \multicolumn{4}{c|}{\cellcolor[HTML]{FFFFFF}Vehicle 3D AP}                                       & \multicolumn{4}{c|}{\cellcolor[HTML]{FFFFFF}Pedestrian 3D AP}                                    & \multicolumn{4}{c}{\cellcolor[HTML]{FFFFFF}Cyclist 3D AP}                                       \\
\rowcolor[HTML]{FFFFFF} 
\cellcolor[HTML]{FFFFFF}                         & \cellcolor[HTML]{FFFFFF}                     & \multicolumn{2}{c}{\cellcolor[HTML]{FFFFFF}L1} & \multicolumn{2}{c|}{\cellcolor[HTML]{FFFFFF}L2} & \multicolumn{2}{c}{\cellcolor[HTML]{FFFFFF}L1} & \multicolumn{2}{c|}{\cellcolor[HTML]{FFFFFF}L2} & \multicolumn{2}{c}{\cellcolor[HTML]{FFFFFF}L1} & \multicolumn{2}{c}{\cellcolor[HTML]{FFFFFF}L2} \\
\rowcolor[HTML]{FFFFFF} 
\multirow{-3}{*}{\cellcolor[HTML]{FFFFFF}Method} & \multirow{-3}{*}{\cellcolor[HTML]{FFFFFF}St} & $IoU_{0.5}$                & $IoU_{0.7}$              & $IoU_{0.5}$                & $IoU_{0.7}$                & $IoU_{0.25}$               & $IoU_{0.5}$               & $IoU_{0.25}$               & $IoU_{0.5}$                & $IoU_{0.25}$            & $IoU_{0.5}$               & $IoU_{0.25}$               & $IoU_{0.5}$              \\ \midrule

\rowcolor[HTML]{FFFFFF}
{  CenterPoint (Boxes Sup)}          & {  ×}                     & {  93.75} & {  72.49} & {  88.13} & {  64.45} & {  89.24} & {  74.28} & {  83.61} & {  66.34} & {  76.67} & {  71.20} & {  75.20} & {  68.58}  \\  \hline 

\rowcolor[HTML]{FFFFFF} 
DBSCAN~\cite{dbscan} + Sem                                     & \textbf{×}                                   & 40.71                  & 9.98                  & 35.60                  & 8.52                   & 67.66                  & 31.20                  & 58.67                   & 26.15                   & 42.42                  & 31.95                 & 40.91                  & 30.77                 \\
\rowcolor[HTML]{FFFFFF} 
DBSCAN + Sem + St                    & \textbf{\checkmark}                                   & 43.07                & 13.09                 & 37.45                  & 11.65                  & 63.99                  & 33.47                  & 56.76                  & 28.16                   & 56.49                  & 32.86                 & 54.47                  & 31.67                 \\
\rowcolor[HTML]{FFFFFF} 
OYSTER~\cite{oyster} + Sem                                     & \textbf{\checkmark}                                   & 45.23                  & 20.36                 & 39.20                   & 17.46                  & 70.45                  & 19.46                   & 61.34                  & 16.40                   & 58.64                  & 37.12                 & 52.43                   & 35.75                  \\ \hline
\rowcolor[HTML]{FFFFFF} 
Seg2Box + Init                          & \textbf{×}                                   & 85.01                  & 56.05                 & 76.46                  & 48.71                  & 81.50                   & \textbf{51.47}        & 73.21                  & \textbf{43.93}         & 60.68                  & 44.08                 & 58.49                  & 42.45                 \\
\rowcolor[HTML]{C0C0C0} 
Seg2Box                                          & \textbf{\checkmark}                                   & \textbf{90.56}         & \textbf{62.28}        & \textbf{83.73}         & \textbf{54.71}         & \textbf{82.35}         & 50.34                 & \textbf{74.31}         & 43.02                  & \textbf{66.56}         & \textbf{46.66}        & \textbf{64.27}         & \textbf{44.96}        \\ \bottomrule
\end{tabular}
\caption{3D object detection results on WOD validation set. St means Self-training. Init means Init-training.}
\label{tab:waymo-exp}
\end{table*}

\begin{table}[ht]
\centering
\setlength{\tabcolsep}{1mm}

\begin{tabular}{c|c|c|ccc}
\toprule
\rowcolor[HTML]{FFFFFF} 
\cellcolor[HTML]{FFFFFF}                         & \cellcolor[HTML]{FFFFFF}                                & \multicolumn{1}{l|}{\cellcolor[HTML]{FFFFFF}}                         & \multicolumn{3}{c}{\cellcolor[HTML]{FFFFFF}Error(↓)}                                       \\
\rowcolor[HTML]{FFFFFF} 
\multirow{-2}{*}{\cellcolor[HTML]{FFFFFF}Method} & \multirow{-2}{*}{\cellcolor[HTML]{FFFFFF}St} & \multicolumn{1}{l|}{\multirow{-2}{*}{\cellcolor[HTML]{FFFFFF}mAP(↑)}} & ATE                          & ASE                          & AOE                          \\ \bottomrule
\rowcolor[HTML]{FFFFFF} 
{  CenterPoint (Boxes Sup)}    & {  ×}                                & {  54.6}                                           & {  0.31} & {  0.26} & {  0.41} \\ \toprule
\rowcolor[HTML]{FFFFFF} 
DBSCAN + Sem                                     & ×                                                       & 28.9                                                                  & 0.42                        & 0.51                         & 1.41                        \\
\rowcolor[HTML]{FFFFFF} 
DBSCAN + Sem + St                     & \checkmark                                                       & 36.0                                                                  & 0.44                        & 0.50                        & 1.51                         \\
\rowcolor[HTML]{FFFFFF} 
OYSTER + Sem                                     & \checkmark                                                       & 33.9                                                                  & 0.46                        & 0.40                          & 1.44                         \\ \midrule
\rowcolor[HTML]{FFFFFF} 
Seg2Box + Init                          & ×                                                       & 44.2                                                                  & 0.37                        & 0.33                        & 1.68                         \\
\rowcolor[HTML]{C0C0C0} 
Seg2Box                                          & \checkmark                                                       & 46.3                                                                  & 0.37                        & 0.32                        & 1.54                        \\ \bottomrule
\end{tabular}
\caption{3D object detection results on nuScenes Dataset validation set. St means Self-training. Init means Init-training.}
\label{tab:nuscenes-exp}
\end{table}

\subsection{Datasets and Metricses}

\subsubsection{Waymo Open Dataset (WOD).}
For bounding box annotation, WOD~\cite{waymo} contains a total of $\sim$158k LiDAR frames for training and $\sim$40k LiDAR frames for validation. For semantic annotation, WOD annotates one frame every 7 frames on the top LiDAR scan, resulting in $\sim$23k frames for training. And, all our experiments are conducted on the frames with semantic annotation and following the official evaluation metrics. 

\subsubsection{NuScenes Dataset.} nuScenes~\cite{nuScenes} is a more challenging dataset since sparser scans and more categories. It contains 1,000 sequences, with 700, 150, and 150 for training, validation, and testing, respectively. 
nuScenes provides new metrics for 3D detection called NDS which comprehensively calculates mAP, Average Translation Error (ATE), Average Scale Error (ASE), Average Orientation Error (AOE), Average Velocity Error (AVE), and Average Attribute Error (AAE). However, we ignore the last two sub-metrics, as they can't be obtained from semantic labels.

\subsection{Implementation Details}
In the pseudo-label generation stage, we used grid size $r = 7$ for Eq.\ref{occ-score} to calculate the Occupancy-Score of the pseudo-label. We used $\lambda_{1} = \lambda_{2} = \lambda_{3} = \frac{1}{3}$ in Eq.\ref{score} for the weights of MSF-Score. We used $\theta_{L} = 0.4$ and $\theta_{H} = 0.8$ in Eq.\ref{loss} to calculate the loss weight of each pseudo-label. In the SGIM-ST stage, We used CenterPoint~\cite{centerpoint} as our based architecture and adopted the implementation of publicly available code from OpenPCDet~\cite{openpcdet2020} for all experiments. We trained both Waymo Open Dataset and nuScenes Dataset for 30 epochs and selected the best validation accuracy epoch as a result. All those experiments were trained on 2 NVIDIA GeForce RTX 3090 GPUs with the ADAM optimizer. For both datasets, we used the same detection ranges as fully supervised methods~\cite{centerpoint}.     

\subsubsection{Baseline.} Seg2Box supervises 3D detectors using only semantic labels, so no previously published baselines exist. We create baselines by pseudo-label generation methods with semantic labels, and we named the modified method as Method-Name + Sem. We first found the baseline DBSCAN + Sem~\cite{dbscan}, since Seg2box follows the idea of density-based spatial clustering to obtain initial pseudo-labels~\cite{oyster, modest}. DBSCAN + Sem + St adds two rounds of self-training. We also assigned semantics to the state-of-the-art unsupervised method OYSTER~\cite{oyster}, indicated as OYSTER + Sem. For a fair comparison, we used the MFF and SCF modules proposed in SGIM-ST for all training, which are intuitive but effective tricks with semantic labels.

\subsection{Main Results}

\subsubsection{Results on WOD.}
The results on the WOD validation set are presented in TABLE \ref{tab:waymo-exp}. The DBSCAN + Sem performs poorly due to the inaccuracy of initial pseudo-labels generated in a single frame. With self-training, DBSCAN + Sem + St outperforms the init-training significantly but is still unsatisfactory due to the poor init-training performance. OYSTER + Sem attempts to refine the pseudo-labels by tracking and refining boxes in long trajectories. 
It still remains a poor performance since the poor tracking performance. 
Our Seg2Box + Init outperforms the best baseline method by 21.8\% in L2 mAP although without self-training. These advancements come from our MFMS-C designs, which generate more accurate pseudo-labels. With SGIM-ST, Seg2Box improves init-training by 7.3\% in 3D AP with an IoU threshold of 0.5 in Vehicle which reaches 95\% of the performance of the fully supervised method. With more strict IoU threshold 0.7, Seg2Box can still reach 85\% performance of fully supervised method.

\begin{table}[ht]

\centering
\setlength{\tabcolsep}{0.75mm}

\begin{tabular}{l|lll|lll}
\toprule
\multicolumn{1}{c|}{}                         & \multicolumn{3}{c|}{3D Recall} & \multicolumn{3}{c}{3D Precision} \\
\multicolumn{1}{c|}{\multirow{-2}{*}{Method}} & $IoU_{0.3}$ & $IoU_{0.5}$ & $IoU_{0.7}$ & $IoU_{0.3}$ & $IoU_{0.5}$ & $IoU_{0.7}$ \\ \midrule
DBSCAN                                        & 45.07    & 24.98    & 10.59    & 41.94     & 23.24     & 9.86     \\
OYSTER                                        & 48.66    & 28.30    & 13.76    & 47.87     & 27.84     & 13.53    \\
\rowcolor[HTML]{C0C0C0} 
Ours                                          & 74.92    & 57.49    & 27.46    & 75.61     & 58.02     & 27.71    \\ \bottomrule
\end{tabular}
\caption{Pseudo-label Comparison Results on WOD. }
\label{tab:pseudo-label-quality}
\end{table}

\begin{figure}[t]
  \centering
   \includegraphics[width=1.0\linewidth]{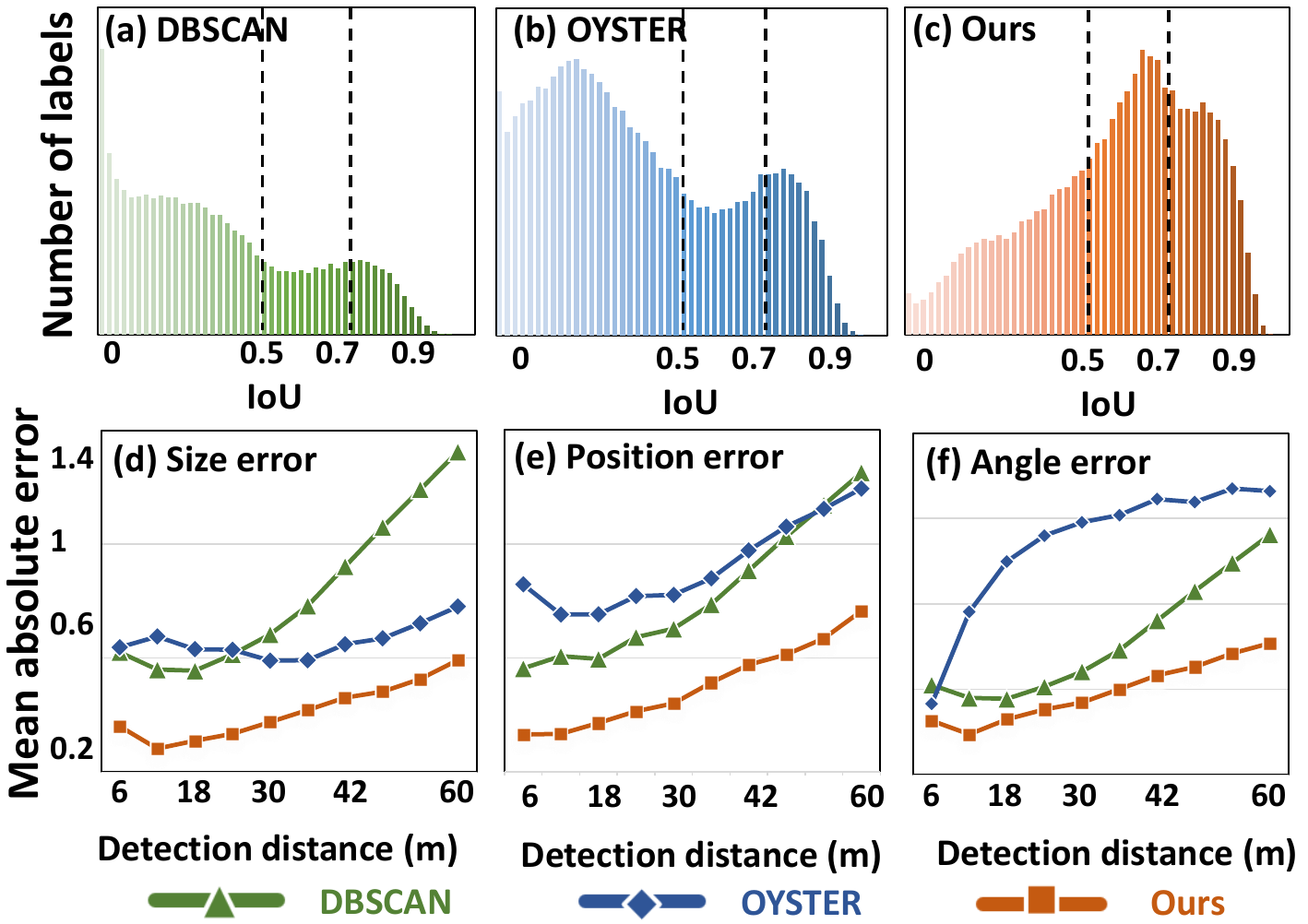}
   \caption{
   (a-c): The IoU distribution between pseudo-labels and ground truth. (d-f): The mean absolute errors (MAEs) for size, position, and angle of pseudo-labels.
   }
   \label{fig:pseudo-labels-error}
\end{figure}

\subsubsection{Results on nuScenes Dataset.}
The results on the nuScenes validation set are presented in TABLE \ref{tab:nuscenes-exp}.  
 Our method demonstrates a 10.3\% improvement in mAP compared to the best baseline detector. In addition, Seg2Box also reduces errors in ATE and ASE. This improvement is attributed to Seg2Box's capacity to generate accurate initial pseudo-labels and its ability to mine and refine pseudo-labels throughout the self-training process continuously.

 \subsection{Pseudo-Label Comparison}
To verify the quality of our pseudo-labels, we calculated the 3D recall and precision on WOD. As shown in TABLE \ref{tab:pseudo-label-quality}, our method outperforms OYSTER~\cite{oyster} with improvements of 13.7\% and 14.18\% in Recall and Precision, respectively, even under a strict 0.7 IoU threshold. To understand the source of our improvements, we evaluated the IoU between the pseudo-labels and the ground truth. The IoU distributions are shown in Fig. \ref{fig:pseudo-labels-error} (a-c). Our method's IoU distribution is more concentrated near 1 with a larger number of high-quality pseudo-labels with IoU scores exceeding 0.7. Additionally, as shown in Fig. \ref{fig:pseudo-labels-error} (d-f), we also evaluated the size error, position error, and angle error of the pseudo-labels. Our method maintains more minor errors even as the detection distance increases. These results confirm that our MFMS-C module significantly reduces label errors and generates higher-quality pseudo-labels.

\subsection{Ablation Study}

\subsubsection{Components Analysis of Seg2Box.} 
To evaluate the individual contributions of Seg2Box, we incrementally added each component and assembled their impact on AP using the WOD validation set in Vehicle. The results are shown in TABLE \ref{tab:abalation_all}. The comparison of the first row and the second row shows that Multi-Frame Clustering (MFC) significantly surpasses Singel-Frame Clustering (SFC) by 26.95\% in AP L2. It indicates that more complete objects in consecutive frames are crucial to generating high-quality pseudo-labels. The third row shows that MSC further enhances performance by 4.3\% in AP L2 since it avoids mistaken instance division to get precise boundaries for pseudo-labels. The last row indicates that SGIM-ST contributes a 6\% increase in AP L2, demonstrating the ability of SGIM-ST to refine pseudo-labels and mine the instances without generating pseudo-labels.

\subsubsection{Component Analysis of SGIM-ST.}
To prove the effect of of SGIM-ST, we incrementally added each component and evaluated their impact on AP using the WOD validation set in Vehicle. The results are shown on TABLE \ref{tab:abalation_st}. The first row presents the result of init-training. The comparison of the first row and the second row shows that our BAF module significantly improves the performance by 5.04\% in AP. It indicates that foreground points without generating pseudo-labels will seriously mislead the training of the model. The third row shows that assembled with MSFLoss, our method further enhances performance by 1.01\% in AP, since it helps inhibit the influence of inaccurate pseudo-labels. With the help of semantic labels, SCF contributes a 2.16\% improvement in AP by filtering the false classification (False positive) of predictions (the fourth row). The last row indicates that STCF contributes a 2.84\% increase in AP, attributed to the refined pseudo-labels of static objects in long-distance.

\begin{table}[t]
\setlength{\tabcolsep}{1.65mm}
\centering
\begin{tabular}{ccll|c|c}
\toprule
\multicolumn{4}{c|}{Seg2Box Components}                                                           & \multirow{2}{*}{3D AP L1} & \multirow{2}{*}{3D AP L2} \\
{SFC} & MFC                  & MSC                  & \multicolumn{1}{c|}{SGIM-ST} &                          &                            \\ \midrule
\checkmark                    & \multicolumn{1}{l}{} &                         &                                             & 20.36                           & 17.46                      \\
                   & \checkmark                    &                         &                                  &  51.33                     & 44.41                         \\
                   & \checkmark                    & \multicolumn{1}{c}{\checkmark}   &                           & 56.05                      & 48.71                      \\
                   & \checkmark                    & \multicolumn{1}{c}{\checkmark}   & \multicolumn{1}{c|}{\checkmark}    & 62.28                      & 54.71                      \\ \bottomrule
\end{tabular}

\caption{Seg2Box component analysis on WOD val. set.  }
    \label{tab:abalation_all}

\end{table}

\begin{table}[t]
\centering
\setlength{\tabcolsep}{1.7mm}


\begin{tabular}{ccll|c|c}
\toprule
\multicolumn{4}{c|}{SGIM-ST Components}                                                           & \multirow{2}{*}{3D AP L1} & \multirow{2}{*}{3D AP L2} \\
BAF                  & MSFLoss              & \multicolumn{1}{c}{SCF} & \multicolumn{1}{c|}{STCF} &                            &                            \\ \midrule
\multicolumn{1}{l}{} & \multicolumn{1}{l}{} &                         &                                                      & 50.37                      & 43.66                      \\
\checkmark                    & \multicolumn{1}{l}{} &                         &                                             & 56.05                      & 48.70                      \\
\checkmark                    & \checkmark                    &                         &                                    & 57.08                          & 49.71                          \\
\checkmark                    & \checkmark                    & \multicolumn{1}{c}{\checkmark}   &                           & 59.51                      & 51.87                      \\
\checkmark                    & \checkmark                    & \multicolumn{1}{c}{\checkmark}   & \multicolumn{1}{c|}{\checkmark}    & 62.28                      & 54.71                      \\ \bottomrule
\end{tabular}

\caption{SGIM-ST component analysis on WOD val. set. }
    \label{tab:abalation_st}
\end{table}

\begin{figure}[t]
  \centering
   \includegraphics[width=1.0\linewidth]{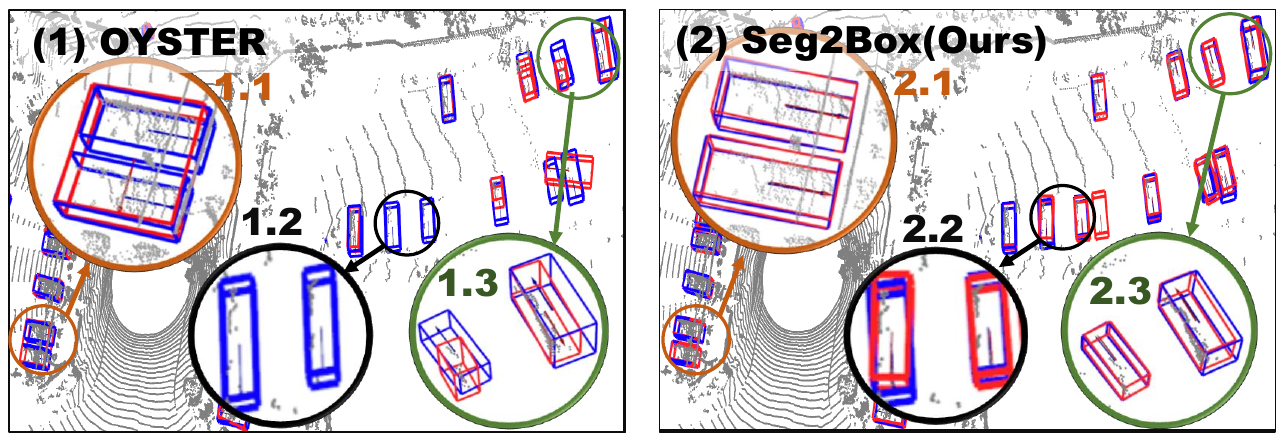}
   \caption{
   Visualization comparison of different detection results on WOD validation set. Blue boxes are the ground truth, and red boxes are the detection results.
   }
   \label{fig:det}
\end{figure}

\subsection{Visualization Comparison of Detection Results}
We visually compare the detection results of different methods in Fig. \ref{fig:det}. OYSTER fails to detect distance, sparse instances (Fig. \ref{fig:det} (1.2)) or detects them with inaccurate sizes and positions (Fig. \ref{fig:det} (1.3)). Additionally, OYSTER also fail to detect the adjacent instances because of the false boundaries of initial pseudo-labels. In contrast, Seg2Box not only identifies these objects but also precisely predicts their sizes and positions. (Fig. \ref{fig:det} (2)).

\section{Conclusion}

This work presents Seg2Box, a novel framework for 3D object detection supervised by semantic labels. It indicates that cross-task supervision between 3D object detection and semantic segmentation is a feasible way to reduce annotation redundancy. First, the MFMS-C module estimates high-quantity pseudo-labels with correct instance divisions, positions and sizes. Furthermore, the SGIM-ST module, is a novel self-training framework designed to refine the pseudo-labels and mine the instances without pseudo-labels iteratively, thereby enhancing detection performance. Experimental results on the Waymo Open Dataset and the nuScenes Dataset demonstrated that Seg2Box outperforms other competitive methods by a large margin. Future work will explore multi-task learning of 3D object detection and 3D semantic segmentation, but using annotation from one task.

\section{Acknowledgments}
This work was supported by the National Natural Science Foundation of China (No.62171393), and the Fundamental Research Funds for the Central Universities (No.20720220064).

\bibliography{main}

\end{document}